\pdfoutput=1

\documentclass[11pt]{article}

\usepackage{acl}

\usepackage{times}
\usepackage{latexsym}

\usepackage[T1]{fontenc}

\usepackage[utf8]{inputenc}

\usepackage{microtype}

\usepackage{multirow}
\usepackage{subcaption}

\usepackage{pgfplots} 
\usetikzlibrary{pgfplots.groupplots}
\usepgfplotslibrary{fillbetween} 

\title{Pre-training Data Quality and Quantity for a Low-Resource Language: New Corpus and BERT Models for Maltese}

\newcommand*{\authormark}[1][*]{\textsuperscript{#1}}
\author{
    Kurt Micallef\authormark[1]\\
    \texttt{kurt.micallef@um.edu.mt}\\\And
    Albert Gatt\authormark[2,3]\\
    \texttt{a.gatt@uu.nl}\\\And
    Marc Tanti\authormark[3]\\
    \texttt{marc.tanti@um.edu.mt}\\\AND
    Lonneke van der Plas\authormark[4,3]\\
    \texttt{lonneke.vanderplas@idiap.ch}\\\And
    Claudia Borg\authormark[1]\\
    \texttt{claudia.borg@um.edu.mt}\\
    \AND 
    {\normalfont\authormark[1]Department of Artificial Intelligence, University of Malta}\\
    {\normalfont\authormark[2]Information and Computing Sciences, Utrecht University}\\
    {\normalfont\authormark[3]Institute of Linguistics and Language Technology, University of Malta}\\
    {\normalfont\authormark[4]Idiap Research Institute}\\
}

\begin{document}
\maketitle

\begin{abstract}
Multilingual language models such as mBERT have seen impressive cross-lingual transfer to a variety of languages, but many languages remain excluded from these models.
In this paper, we analyse the effect of pre-training with monolingual data for a low-resource language that is not included in mBERT -- Maltese -- with a range of pre-training set ups.
We conduct evaluations with the newly pre-trained models on three morphosyntactic tasks -- dependency parsing, part-of-speech tagging, and named-entity recognition -- and one semantic classification task -- sentiment analysis.
We also present a newly created corpus for Maltese, and determine the effect that the pre-training data size and domain have on the downstream performance.
Our results show that using a mixture of pre-training domains is often superior to using Wikipedia text only.
We also find that  a fraction of this corpus is enough to make significant leaps in performance over Wikipedia-trained models.
We pre-train and compare two models on the new corpus: a monolingual BERT model trained from scratch (BERTu), and a further pre-trained multilingual BERT (mBERTu).
The models achieve state-of-the-art performance on these tasks, despite the new corpus being considerably smaller than typically used corpora for high-resourced languages.
On average, BERTu outperforms or performs competitively with mBERTu, and the largest gains are observed for higher-level tasks.
\end{abstract}

\section{Introduction}\label{section:introduction}

Language Models have become a core component in many Natural Language Processing (NLP) tasks.
These models are typically pre-trained on unlabelled texts, and then further fine-tuned using labelled data relevant to the target task.
Transformer-based \cite{NIPS2017_3f5ee243} contextual models such as BERT \cite{devlin-etal-2019-bert} have gained success since the fine-tuning step is relatively inexpensive, while attaining state-of-the-art results in various syntactic and semantic tasks.

While the bulk of work with the BERT family of models focuses on English, there have been some monolingual models developed for other languages as well (\citealp{martin-etal-2020-camembert, polignano2019alberto, antoun-etal-2020-arabert, BERTje, FinBERT, agerri-etal-2020-give}; inter alia).
These monolingual models have been trained on large volumes of data, typically amounting to billions of tokens. 
In contrast, it is challenging to find publicly available corpora of this size for low-resource languages.
The evaluation benchmarks for downstream tasks on these languages are also limited, and tend to be dominated by low-level structural tagging tasks.

To counteract the lack of large volumes of monolingual corpora for low-resource languages, a number of multilingual models have been released, such as mBERT \cite{devlin-etal-2019-bert} and XLM-R \cite{conneau-etal-2020-unsupervised}.
These multilingual models were pre-trained on more than one language at a time by combining corpora from different languages, usually sourced from Wikipedia.
Several works have demonstrated the efficacy of these multilingual models, especially for languages without a language-specific model \cite{kondratyuk-straka-2019-75, wu-dredze-2019-beto}.
Benchmark results have improved for many languages by leveraging cross-linguistic features learnt by these multilingual models \cite{conneau-etal-2020-unsupervised}.

However, the gains with multilingual models may vary depending on the language being considered.
The ``curse of multilinguality'' limits the language-specific features that these models can learn, since the limited model capacity has to be shared between multiple languages \cite{conneau-etal-2020-unsupervised}.
Models such as mBERT use WordPiece tokenisation \cite{johnson-etal-2017-googles}, which splits words into various sub-tokens, thereby reducing the number of unknown tokens.
However, the vocabulary representations for multilingual models tend to be sub-optimal for specific languages, because words tend to be split into a higher number of sub-tokens \cite{rust-etal-2021-good}.
Moreover, these models may still be biased in favour of over-representing sub-tokens common to a certain subset of languages over others.
Due to the data imbalance across languages, lower-resourced languages tend to be disadvantaged, as there is relatively less pre-training data available compared to the other languages considered in the multilingual model \cite{wu-dredze-2020-languages}. 

Apart from the tension between languages in a multilingual model, other factors are at play as well.
Most prominently, many languages are never seen by these multilingual models \cite{muller-etal-2021-unseen}, since these are typically trained on the largest-available corpora (e.g. mBERT was pre-trained on the 104 languages with the greatest Wikipedia presence).
Such criteria exclude many of the world's languages, including Maltese, the focus of this paper.
This issue is exacerbated even further when the language uses a script which is either different to its closely related languages \cite{muller-etal-2021-unseen}, or which is never seen during pre-training, thereby encoding most of the input with out-of-vocabulary tokens \cite{pfeiffer-etal-2021-unks}.
In fact, \citet{muller-etal-2021-unseen} show that the language transfer capability of a multilingual model to an unseen language is dependent on the degree to which the target language is related to languages already included in the multilingual model.

In this work we focus on the Maltese language, an official EU language spoken primarily in Malta and in some small communities around the world \cite{Brincat2011}. 
It is the only Semitic language written exclusively with a Latin script, containing a few additional characters with diacritic marks (ċ, ġ, \hwithstroke, ż).
The language also has strong influences from Romance languages such as Italian, as well as English.
The Semitic influence is largely exhibited in the grammatical structure through complex morphological characteristics, whilst the non-Semitic aspect is predominantly observed in its vocabulary, with extensive lexical borrowing from Italian and English.

In the context of NLP, Maltese is a low-resource language~\cite{Rosner2022} and is not part of the languages covered by either mBERT or XLM-R.
\citet{muller-etal-2021-unseen} find that mBERT underperforms non-contextual baselines on Maltese, but benefits when pre-trained further on raw Maltese data.
Similarly, \citet{chau-etal-2020-parsing} further pre-train mBERT but impute the 99 unused tokens present in the model with language specific tokens, yielding  better results. This confirms previous findings by \citet{wang-etal-2020-extending}, who also extend mBERT's vocabulary to accommodate unseen languages, but do so by extending the vocabulary and model dimensionality, hence increasing its footprint.

Motivated by the limitations of existing multilingual models and the deficiency of publicly available corpora for Maltese, we set out to pre-train a new monolingual language model for Maltese and compare it to the alternative strategy of further pre-training an existing multilingual model.
We study, in particular, the impact that the pre-training data size and domain has on the performance in downstream tasks.
The main contributions of this work are as follows:
\begin{enumerate}
    \item We develop a new corpus of Maltese text.
    \item Using this new data, language models for Maltese are pre-trained.
    \item We compare the newly pre-trained models and find that both models improve the state-of-the-art on three structural tagging tasks -- dependency parsing, part-of-speech tagging, and named-entity recognition -- and one semantic classification task -- sentiment analysis.
    \item We demonstrate that in a low-resource setting, pre-training using text from varied domains is often superior to solely using Wikipedia, and that matching the domain to target task is beneficial when this is available.
    \item We also provide an analysis on the effects of the pre-training size, shedding new light on how much pre-training data is needed to attain significant improvements in performance.
\end{enumerate}

We make this new corpus, the newly pre-trained language models, and the code publicly available\footnote{The corpus and the language models are available at the Hugging Face Hub at \url{https://huggingface.co/datasets/MLRS/korpus_malti}, \url{https://huggingface.co/MLRS/BERTu}, and \url{https://huggingface.co/MLRS/mBERTu}. The code is available at \url{https://github.com/MLRS/BERTu}.}.

\section{Corpus}\label{section:corpus}

In this work, we build a new unlabelled text corpus, which we call the \textbf{Korpus Malti v4.0 (KM)}.
This builds on and extends an existing corpus, Korpus Malti v3.0\footnote{See: \url{https://mlrs.research.um.edu.mt}}, which is approximately half the size.

Rather than scraping the web randomly for Maltese text, we collect text data from specific sources, including both online and offline.
Although this does incur additional effort in data collection, and results in a smaller dataset compared to large-scale web-scraping initiatives, it has the benefit of resulting in a less noisy dataset, while offering greater control over sources.
For comparison, the Maltese portion of the OSCAR data \cite{OrtizSuarezSagotRomary2019}, which is sourced entirely from the web, contains texts which, to a native speaker, suggest that they are automatically generated through the use of a low-quality machine translation system, a common pitfall of web-scraping for low-resource languages \cite{10.1162/tacl_a_00447}.
We also expect to find a small proportion of code-switched texts, as this is a pre-dominant phenomenon for Maltese in domains such as social media or transcribed speech.
In addition, the data is separated into different domains, and the source for each document is available as part of the metadata.
This allows data users to select data subsets which are more appropriate for their particular use-case, such as domain-adaptive pre-training (\citealp{BioBERT, gururangan-etal-2020-dont}; inter alia), whilst enabling tracing back to the original source, or omission in case unforeseen ethical or privacy issues come to light.
In short, the goal was to build a good quality training dataset, while avoiding at least some of the pitfalls identified with opportunistic, web-scale data initiatives \cite{bender-2021-parrots, rogers-2021-changing}.

Data is collected from a variety of sources, including online news sources, legal texts, transcripts of speeches and debates, blogs, Wikipedia, etc.
Before texts are included in the corpus, we filter non-Maltese sentences using language identification using LiD \cite{lui-baldwin-2014-accurate}, and perform de-duplication using Onion \cite{pomikalek2011removing}.

The resulting data, split into 19 different domains, is summarised by Table~\ref{table:data}.

\begin{table*}
    \centering
    \begin{tabular}{|l|r|r|r|r|}
        \hline
        \multicolumn{1}{|l|}{\textbf{data subset}} & \multicolumn{1}{c|}{\textbf{documents}} & \multicolumn{1}{c|}{\textbf{sentences}} & \multicolumn{1}{c|}{\textbf{tokens}} & \multicolumn{1}{c|}{\textbf{size}} \\
        \hline
        belles\_lettres & $195$ & $299\,762$ & $4\,454\,906$ & $21.82$MB \\
        blogs & $25\,436$ & $807\,628$ & $14\,562\,039$ & $74.45$MB \\
        comics & $62$ & $2\,413$ & $44\,768$ & $233.22$KB \\
        court & $2\,663$ & $694\,227$ & $11\,881\,638$ & $61.91$MB \\
        eu\_docs & $2\,974$ & $5\,099\,564$ & $135\,811\,945$ & $773.25$MB \\
        government\_gazette & $2\,974$ & $1\,881\,034$ & $39\,771\,556$ & $203.61$MB \\
        gov\_docs & $272$ & $120\,209$ & $1\,900\,842$ & $10.79$MB \\
        law\_eu & $71$ & $4\,433\,235$ & $98\,582\,031$ & $541.13$MB \\
        law\_mt & $2\,596$ & $401\,118$ & $7\,631\,651$ & $38.84$MB \\
        legal & $3$ & $4\,784$ & $83\,581$ & $490.67$MB \\
        nonfiction & $2\,177$ & $208\,763$ & $3\,902\,436$ & $20.01$MB \\
        parliament & $6\,198$ & $3\,935\,906$ & $82\,294\,520$ & $433.09$MB \\
        press\_eu & $5\,483$ & $413\,317$ & $9\,774\,919$ & $55.73$MB \\
        press\_mt & $46\,782$ & $713\,886$ & $17\,679\,904$ & $93.15$MB \\
        speeches & $62$ & $2\,067$ & $51\,259$ & $286.63$MB \\
        theses & $19$ & $11\,545$ & $310\,243$ & $1.63$MB \\
        umlib\_oar & $11\,688$ & $963\,606$ & $21\,235\,949$ & $106.11$MB \\
        web\_general & $2$ & $685\,873$ & $14\,741\,525$ & $75.22$MB \\
        wiki & $3\,469$ & $79\,134$ & $1\,885\,661$ & $9.73$MB \\
        \hline
        all & $131\,429$ & $20\,758\,071$ & $466\,601\,373$ & $2.52$GB \\
        \hline
    \end{tabular}
    \caption{Korpus Malti v4.0 corpus distribution. \textit{belles\_lettres} is largely composed of literary works; the \textit{government\_gazzette} consists of text from the official newsletter of the Maltese government; \textit{umlib\_oar} is a miscellaneous collection of previously published non-fiction texts, available in the public domain via the University of Malta Library Open Access Repository.}\label{table:data}
\end{table*}

To the best of our knowledge, there is no corpus of this size available for Maltese. 
We also note that this data is a significant increase over Wikipedia data, which is what is usually available and used in low-resource scenarios.
The Wikipedia data makes up less than 1\% of the entire corpus in terms of both tokens and sentences.

Despite this substantial increase in data, we emphasise that a corpus of under 500M tokens is still substantially smaller than is typically used for higher-resourced languages.
For example, \citet{devlin-etal-2019-bert} pre-train BERT using a combined corpus of 3.3B words for English (approximately 16GB).
Larger models have since exceeded these pre-training sizes by a wide margin -- for example, RoBERTa is pre-trained on 161GB of text \cite{liu-etal-2019-roberta}.
Monolingual models for languages other than English, typically use smaller corpora than English models, but their size is still significantly larger than ours -- for example AraBERT was pre-trained on a corpus of 24GB \cite{antoun-etal-2020-arabert} and BERTje was pre-trained on a corpus of 12GB \cite{BERTje}.

\section{Language Models}\label{section:models}

Using this new corpus, two new language models are pre-trained for Maltese: a monolingual model (\textbf{BERTu}) and a multilingual model (\textbf{mBERTu}).
In both cases, pre-training is performed using the Masked Language Modelling Objective (MLM) only, since the Next Sentence Prediction (NSP) objective was found to be detrimental to downstream performance \cite{joshi-etal-2020-spanbert, liu-etal-2019-roberta}.
Other than that, pre-training largely follows the pre-training setup of BERT \cite{devlin-etal-2019-bert}.
This allows for a better comparison with already available models.
The pre-training data from all domains is combined, shuffled, and split into 85\% and 15\% for training and validation sets respectively.

\paragraph{BERTu}
We pre-train a monolingual BERT model from scratch on the new unlabelled data, using the BERT\textsubscript{\textsc{BASE}} architecture with 12 transformer layers, a hidden size of 768, and 12 attention heads.
The vocabulary is initialised with 52K tokens.
Pre-training is done across 1M steps, with a sequence length of 128 for the first 90\% of the steps and a sequence length of 512 for the remaining 10\% steps.
A batch size of 512 is used, which amounts to approximately 30 epochs in total, and a warmup of 1\% of the total number of steps.
We use mixed-precision training to ease memory requirements.
Training was performed on 8 A100 GPUs for the first 90\% steps and 16 A100 GPUs for the remaining 10\% steps, taking approximately 53 hours.

\paragraph{mBERTu}
Similar to \citet{chau-etal-2020-parsing} and \citet{muller-etal-2021-unseen} we also pre-train mBERT further on Maltese.
Since the embedding weights are not randomly initialised, as is the case for the monolingual model, we follow \citet{rust-etal-2021-good} and pre-train for 250K steps.
A sequence length of 512 is used throughout, keeping the rest of the hyper-parameters the same as the monolingual pre-training.
To better fit the Maltese language, the mBERT vocabulary is augmented with Maltese tokens following the procedure from \citet{chau-etal-2020-parsing}, by replacing the unused tokens reserved in the original vocabulary.
Specifically, we train a tokeniser with a vocabulary size of $5\,000$ tokens on the data and choose the set of 99 tokens which reduce the number of \texttt{[UNK]} tokens the most in the target data.
Training was performed on 32 A100 GPUs, and took around 46 hours to complete.

\section{Evaluation}\label{section:evaluation}

An evaluation for the language models described in Section~\ref{section:models} is presented here.
\textbf{mBERT} without any additional pre-training is used as one of the baselines.
In addition, we pre-train two language models on the Maltese Wikipedia data as additional baselines.
This allows us to analyse the limitations that could be faced when following the common practice of using Wikipedia data, for the specific case of low-resource languages with a comparatively small Wikipedia footprint.

Following the same setup of the main models, a monolingual model (\textbf{BERTu Wiki}) and a multilingual model (\textbf{mBERTu Wiki}) are pre-trained.
The same hyper-parameters described in Section~\ref{section:models} are used, but the batch size and number of steps are decreased to prevent overfitting due to the smaller data size.
To this end, the batch size is set to 64 and the total number of steps set to $30\,500$ and $7\,600$ steps for the monolingual and multilingual models, respectively.
This was deemed appropriate since it would amount to the same number of epochs as the models pre-trained on the entire corpus.

\subsection{Tasks}

The language models are fine-tuned on the following downstream tasks.
A summary of the datasets and fine-tuning architectures used is given below.

\paragraph{Dependency Parsing (DP)}
The Maltese Universal Dependencies Treebank (MUDT) \cite{MUDT} is used for this task using the provided training, validation, and testing splits.
The data is composed of $2\,074$ human-annotated sentences from 4 different high-level domains.
Similar to \citet{chau-etal-2020-parsing}, \citet{muller-etal-2021-unseen}, and \citet{chau-smith-2021-specializing}, we use a Biaffine graph-based prediction layer \cite{dozat-2019-biaffine} and use the Labelled Attachment Score (LAS) as the main evaluation metric, but also report the Unlabelled Attachment Score (UAS).

\paragraph{Part-of-Speech Tagging (POS)}
The MLRS POS data \cite{gatt-ceplo-2013-korpus}, is used for this task.
This data is composed of $6\,167$ human-annotated sentences -- $426$ of which overlap with the MUDT data \cite{MUDT} -- and are stratified into 8 domains.
We combine the data from the different domains, shuffle it, and split the data into 80\%, 10\%, and 10\% for training, validation, and testing sets, respectively.
The annotations are language-specific tags (using the XPOS scheme) and we follow the tag mapping in MUDT \cite{MUDT} to also produce tags in the Universal Part of Speech tagset (UPOS).
To evaluate tagging with these two tagsets, we use a linear layer, and use accuracy as the evaluation metric.

\paragraph{Named-Entity Recognition (NER)}
The Maltese annotations for the WikiAnn data \cite{pan-etal-2017-cross} are used for this task, using the data splits from \citet{rahimi-etal-2019-massively}.
The data is made up of $300$ sentences derived from Wikipedia.
Following \citet{chau-smith-2021-specializing}, a Conditional Random Field layer is used for this task, and we use F1 as the evaluation metric.

\paragraph{Sentiment Analysis (SA)}
We use the Maltese sentiment analysis dataset by \citet{martinez-garcia-etal-2021-evaluating}, which is a collection of $815$ sentences, using the provided training, validation, and testing splits.
The texts in this data originate from comments on news articles and social media posts, and are a combination of two datasets from \citet{cortis-davis-2019-social} and \citet{dingli-sant-2016-sentiment}.
A linear prediction layer is used, and we use the macro-averaged F1 score as the evaluation metric.

We largely use the hyper-parameters from \citet{chau-smith-2021-specializing}, but optimise the learning rate, batch size, and dropout on the validation set of each task.
Table~\ref{table:hyperparameters} shows the chosen hyper-parameters.
Fine-tuning is performed for at most 200 epochs, with an early stopping of 20 epochs  on the validation set.

\begin{table}
    \centering
    \begin{tabular}{|l|cccc|}
        \hline
        Name & DP & POS & NER & SA \\
        \hline
        Learning Rate & 5e-4 & 5e-4 & 5e-4 & 1e-4 \\
        Batch Size & 128 & 128 & 64 & 32 \\
        Dropout & 0.3 & 0.3 & 0.2 & 0.5 \\
        \hline
    \end{tabular}
    \caption{Fine-tuning hyper-parameters}
    \label{table:hyperparameters}
\end{table}

\subsection{Results}\label{section:results}

The results on all tasks are summarised in Table~\ref{table:results}.
Consistent with the results reported by \citet{muller-etal-2021-unseen} and \citet{chau-smith-2021-specializing}, BERTu Wiki generally underperforms mBERT, and mBERTu Wiki performs better than mBERT.
Whilst they show this for Dependency Parsing, Part-of-Speech tagging, and Named Entity Recognition, we demonstrate that this also holds for Sentiment Analysis.

Our baseline results diverge slightly from previous results on the Named-Entity Recognition task, where BERTu Wiki performs slightly better than mBERT.
We suspect that this is due to a slightly different pre-training setup than that used by \citet{muller-etal-2021-unseen} and \citet{chau-smith-2021-specializing}\footnote{\citet{muller-etal-2021-unseen} pre-train for at most 10 epochs whilst \citet{chau-smith-2021-specializing} pre-train for at most 20 epochs (choosing the best performing model on based on the validation set). Both use a smaller-sized BERT architecture with 6 layers and pre-train with a maximum sequence length of 128.}, but we analyse this further in Section~\ref{section:analysis-domains}.
However, these results remain consistent with regards to BERTu Wiki not performing as well as mBERTu Wiki.

\begin{table*}
    \centering
    \begin{subtable}{0.55\linewidth}
        \centering
        \begin{tabular}{|ll|rr|}
            \hline
            Data & Model & \multicolumn{1}{c}{UAS} & \multicolumn{1}{c|}{LAS} \\
            \hline\hline
            \multirow{2}{*}{Wiki} & BERTu & 80.95 \small{\textpm~0.25} & 74.16 \small{\textpm~0.20} \\
            & mBERTu & 88.74 \small{\textpm~0.11} & 82.59 \small{\textpm~0.19} \\
            \hline
            \multirow{1}{*}{N/A} & mBERT & 84.83 \small{\textpm~0.31} & 77.22 \small{\textpm~0.34} \\
            \hline
            \multirow{2}{*}{KM} & BERTu & ~\textbf{92.31 \small{\textpm~0.15}} & ~\textbf{88.14 \small{\textpm~0.21}} \\
            & mBERTu & *92.10 \small{\textpm~0.14} & *87.87 \small{\textpm~0.18} \\
            \hline
        \end{tabular}
        \caption{Dependency Parsing}\label{table:results-dp}
    \end{subtable}
    \begin{subtable}{0.4\linewidth}
        \centering
        \begin{tabular}{|ll|r|}
            \hline
            Data & Model & \multicolumn{1}{c|}{span F1} \\
            \hline\hline
            \multirow{2}{*}{Wiki} & BERTu & 67.96 \small{\textpm~2.20} \\
            & mBERTu & *85.01 \small{\textpm~2.92} \\
            \hline
            \multirow{1}{*}{N/A} & mBERT & 65.41 \small{\textpm~2.06} \\
            \hline
            \multirow{2}{*}{KM} & BERTu & ~\textbf{86.77 \small{\textpm~3.55}} \\
            & mBERTu & *86.60 \small{\textpm~2.49} \\
            \hline
        \end{tabular}
        \caption{Named Entity Recognition}\label{table:results-ner}
    \end{subtable}
    \begin{subtable}{0.55\linewidth}
        \centering
        \begin{tabular}{|ll|rr|}
            \hline
            Data & Model & \multicolumn{1}{c}{UPOS} & \multicolumn{1}{c|}{XPOS} \\
            \hline\hline
            \multirow{2}{*}{Wiki} & BERTu & 97.27 \small{\textpm~0.11} & 97.01 \small{\textpm~0.07} \\
            & mBERTu & 97.95 \small{\textpm~0.13} & 97.83 \small{\textpm~0.08} \\
            \hline
            \multirow{1}{*}{N/A} & mBERT & 97.26 \small{\textpm~0.15} & 97.20 \small{\textpm~0.14} \\
            \hline
            \multirow{2}{*}{KM} & BERTu & 98.58 \small{\textpm~0.02} & *98.54 \small{\textpm~0.03} \\
            & mBERTu & ~\textbf{98.66 \small{\textpm~0.03}} & ~\textbf{98.58 \small{\textpm~0.04}} \\
            \hline
        \end{tabular}
        \caption{Part-of-Speech tagging}\label{table:results-pos}
    \end{subtable}
    \begin{subtable}{0.4\linewidth}
        \centering
        \begin{tabular}{|ll|r|}
            \hline
            Data & Model & \multicolumn{1}{c|}{macro-F1} \\
            \hline\hline
            \multirow{2}{*}{Wiki} & BERTu & 53.95 \small{\textpm~2.70} \\
            & mBERTu & 56.05 \small{\textpm~3.24} \\
            \hline
            \multirow{1}{*}{N/A} & mBERT & 55.99 \small{\textpm~3.63} \\
            \hline
            \multirow{2}{*}{KM} & BERTu & ~\textbf{78.96 \small{\textpm~1.95}} \\
            & mBERTu & *76.79 \small{\textpm~1.79} \\
            \hline
        \end{tabular}
        \caption{Sentiment Analysis}\label{table:results-sa}
    \end{subtable}
    \caption{
        Experimental results, grouped by the underlying language model and additional pre-training data used.
        All figures shown are the mean and standard deviations over 5 runs with different random seeds.
        The best performing models for each metric are \textbf{bolded}.
        Values marked with * are not found to be significantly worse than the best model (using a 1-tailed \textit{t}-test with a \textit{p}-value = 0.05 with Bonferroni correction).
    }\label{table:results}
\end{table*}

Both language models pre-trained with the KM data perform significantly better than all the other baselines, on all tasks except for Named-Entity Recognition, where the trend is similar, but does not reach statistical significance.
This underlines the value of this new corpus.
When compared to the Wikipedia language models, the most noticeable improvements can be seen between the BERTu models, across all tasks.
A more detailed analysis on this is presented in Section~\ref{section:analysis-size}, but intuitively this finding makes sense since mBERTu models are exposed to significantly more data, making them less specific to Maltese.

The gap in performance between the BERTu and mBERTu models is much less for the KM pre-trained models than it is for the Wikipedia pre-trained models.
In fact, on average, the BERTu KM model performs better than the mBERTu KM model on all tasks except Part-of-Speech tagging.
For Part-of-Speech tagging we also note that the baseline results are already quite high, probably due to the relatively larger labelled data, which may partially mask the effects of the KM models.

Overall, Sentiment Analysis is the task where the most gains are made with respect to the baselines.
The KM-trained models are over 20 F1 points higher than the best performing baseline.
This finding provides evidence that, unlike syntactic tasks, where structural information could potentially be shared across related clusters of languages, semantic tasks such as Sentiment Analysis will benefit much more from language-specific embeddings.

\section{Analysis}\label{section:analysis}

In this section we build on the results presented in Section~\ref{section:results}, and analyse the effect that the pre-training data has on performance on the downstream tasks.

\subsection{Data Domain}\label{section:analysis-domains}

In this subsection we take advantage of the fact that the data is stratified by domains.
Here, we analyse the impact of pre-training using data from different domains, compared to a single domain, namely Wikipedia, which is commonly used in multilingual models and low-resource settings.
For this purpose, we consider the BERTu Wiki and mBERTu Wiki baselines from Section~\ref{section:evaluation} as single-domain models.
We compare them to language models pre-trained with the same amount of data but from different domains, which are referred to as ``Mixed'' in this discussion.

We determine the size using the number of sentences, since this would directly effect the number of epochs and allows us to keep an identical pre-training setup as the Wikipedia-trained models.
Since the Maltese Wikipedia data is composed of $79\,134$ sentences, the Mixed language models are also pre-trained with the same amount of sentences, split into training and validation sets as the Wikipedia models.
A comparison of the downstream task performance for these language models is plotted in Figure~\ref{figure:domain}.

\newcommand{\domainplot}[3]{
    \nextgroupplot[
        title = {#3},
        xlabel = {Pre-training Data Domain},
        ylabel = {#2},
        legend style = {
            legend columns = -1,
            align = center,
            column sep = 10pt,
        },
        legend to name = {legend:domain},
        symbolic x coords = {Wiki, Mixed},
        xtick distance = 1,
    ]
            
    \addplot+[
        error bars/.cd,
        y dir = both,
        y explicit,
    ] table[
        x = {data},
        y expr = \thisrow{#1_m} * 100,
        y error plus expr = \thisrow{#1_std} * 100,
        y error minus expr = \thisrow{#1_std} * 100,
    ] {bert_mt_small.dat};
    \addlegendentry{BERTu}
            
    \addplot+[
        error bars/.cd,
        y dir = both,
        y explicit,
    ] table[
        x = {data},
        y expr = \thisrow{#1_m} * 100,
        y error plus expr = \thisrow{#1_std} * 100,
        y error minus expr = \thisrow{#1_std} * 100,
    ] {mbert_mt_small.dat};
    \addlegendentry{mBERTu}
            
    \addplot+[
        black,
        dashed,
        mark = none,
    ] table[
        y expr = \thisrow{#1_m} * 100,
    ] {mbert_small.dat};
    \addlegendentry{mBERT}
    
    \addplot[
        name path = upper,
        draw = none,
    ] table[
        x = {data},
        y expr = (\thisrow{#1_m} + \thisrow{#1_std}) * 100,
    ] {mbert_small.dat};
    \addplot[
        name path = lower,
        draw = none,
    ] table[
        x = {data},
        y expr = (\thisrow{#1_m} - \thisrow{#1_std}) * 100,
    ] {mbert_small.dat};
    \addplot[
        fill = black,
        opacity = 0.1,
    ] fill between[of=upper and lower];
}
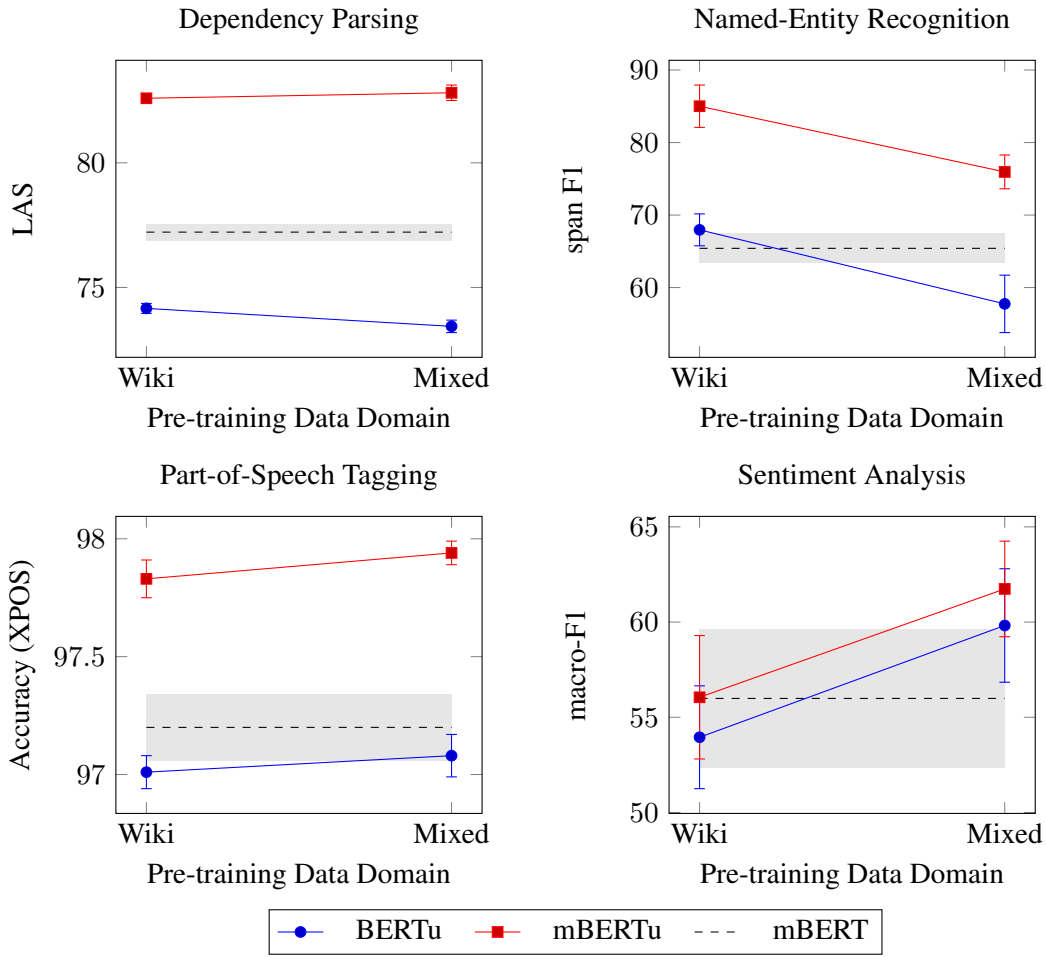
\begin{figure*}[t]
    \centering
    \begin{tikzpicture}
        \begin{groupplot}[
            group style={
                group size = 2 by 2,
                horizontal sep = 70pt,
                vertical sep = 60pt,
            },
            width = 0.4\textwidth,
        ]
            \domainplot{LAS}{LAS}{Dependency Parsing}
            \coordinate (column1) at (rel axis cs:0,1);
            \domainplot{NER}{span F1}{Named-Entity Recognition}
            \coordinate (column2) at (rel axis cs:1,1);
            \domainplot{XPOS}{Accuracy (XPOS)}{Part-of-Speech Tagging}
            \domainplot{SA}{macro-F1}{Sentiment Analysis}
        \end{groupplot}
        \coordinate (centre) at ($(column1)!.5!(column2)$);
        \node[below] at (centre |- current bounding box.south){\pgfplotslegendfromname{legend:domain}};
    \end{tikzpicture}
    \caption{
        Downstream task performance with different pre-training data domains.
        All values are the mean over 5 runs with different random seeds.
        The standard deviation is represented by the corresponding error bars and shaded area.
    }\label{figure:domain}
\end{figure*}

At this small scale of data, both Wiki and Mixed mBERTu models consistently perform better than the mBERT models, owing to the multilingual representation power of these models.
The Mixed models perform better than their Wikipedia counterparts on Part-of-Speech tagging and Sentiment Analysis.
On Dependency Parsing, there is a slight improvement on the mBERTu model but a slight degradation on the BERTu model.

For Named-Entity Recognition, the Wikipedia models perform better than the Mixed ones.
Since the dataset for this task originates from Wikipedia as well, 
this indicates that matching the pre-training data domain to the target domain boosts performance, supporting the findings by \citet{gururangan-etal-2020-dont}.
In fact, mBERTu Wiki surpasses mBERT, and proves to be a competitive baseline, as shown in Table~\ref{table:results-ner}.
On the other hand, mBERTu Mixed performs worse than mBERT.

The opposite is true for Sentiment Analysis, as BERTu Mixed turns out to be a more competitive baseline than mBERT.
The improvement is so pronounced for this task that the BERTu Mixed model not only performs better than the BERTu Wiki counterpart, but also better than mBERTu Wiki.
Even though this dataset contains texts exhibiting stylistic features expected in social media text, a mixture of domains is helpful, probably since Wikipedia texts tend to be quite structured and neutral in terms of the writing style and tone.
The results on sentiment analysis suggest that pre-training on a diversity of domains contribute to more effective learning of features relevant to discourse semantic tasks, compared to tasks involving morpho-syntactic tagging.
We leave further investigation of this, on a broader range of semantically-oriented tasks, for future work.

Overall, these results emphasise the importance of having pre-training data from sources close to the target data, even for low-resource settings.

\subsection{Data Size}\label{section:analysis-size}

From Table~\ref{table:results}, it is clear that the KM corpus translates to better performance on downstream tasks, regardless of whether a monolingual or a multilingual model is used.
To better understand the relationship between the data size and performance, we pre-train several language models with varying data sizes.

\newcommand{\dataplot}[3]{
    \nextgroupplot[
        title = {#3},
        xlabel = {Pre-training Data Proportion (\%)},
        ylabel = {#2},
        xmin = 0, xmax = 100,
        legend style = {
            legend columns = -1,
            align = center,
            column sep = 10pt,
        },
        legend to name = {legend:size},
    ]
    
    \addplot table[
        x expr = \thisrow{data} * 100,
        y expr = \thisrow{#1_m} * 100,
    ] {bert_mt.dat};
    \addlegendentry{BERTu}
    
    \addplot table[
        x expr = \thisrow{data} * 100,
        y expr = \thisrow{#1_m} * 100,
    ] {mbert_mt.dat};
    \addlegendentry{mBERTu}
    
    \addplot[
        black,
        dashed,
        mark = none,
    ] table[
        x expr = \thisrow{data} * 100,
        y expr = \thisrow{#1_m} * 100,
    ] {mbert.dat};
    \addlegendentry{mBERT}
    
    \addplot[
        name path = upper,
        draw = none
    ] table[
        x expr = \thisrow{data} * 100,
        y expr = (\thisrow{#1_m} + \thisrow{#1_std}) * 100,
    ] {bert_mt.dat};
    \addplot[
        name path = lower,
        draw = none,
    ] table[
        x expr = \thisrow{data} * 100,
        y expr = (\thisrow{#1_m} - \thisrow{#1_std}) * 100,
    ] {bert_mt.dat};
    \addplot[
        fill = blue,
        opacity = 0.1,
    ] fill between[of=upper and lower];
    
    \addplot[
        name path = upper,
        draw = none,
    ] table[
        x expr = \thisrow{data} * 100,
        y expr = (\thisrow{#1_m} + \thisrow{#1_std}) * 100,
    ] {mbert_mt.dat};
    \addplot[
        name path = lower,
        draw = none,
    ] table[
        x expr = \thisrow{data} * 100,
        y expr = (\thisrow{#1_m} - \thisrow{#1_std}) * 100,
    ] {mbert_mt.dat};
    \addplot[
        fill = red,
        opacity = 0.1,
    ] fill between[of=upper and lower];
    
    \addplot[
        name path = upper,
        draw = none,
    ] table[
        x expr = \thisrow{data} * 100,
        y expr = (\thisrow{#1_m} + \thisrow{#1_std}) * 100,
    ] {mbert.dat};
    \addplot[
        name path = lower,
        draw = none,
    ] table[
        x expr = \thisrow{data} * 100,
        y expr = (\thisrow{#1_m} - \thisrow{#1_std}) * 100,
    ] {mbert.dat};
    \addplot[
        fill = black,
        opacity = 0.1,
    ] fill between[of=upper and lower];
}
\begin{figure*}[t]
    \centering
    \begin{tikzpicture}
        \begin{groupplot}[
            group style={
                group size = 2 by 2,
                horizontal sep = 70pt,
                vertical sep = 60pt,
            },
            width = 0.4\textwidth,
        ]
            \dataplot{LAS}{LAS}{Dependency Parsing}
            \coordinate (column1) at (rel axis cs:0,1);
            \dataplot{NER}{span F1}{Named-Entity Recognition}
            \coordinate (column2) at (rel axis cs:1,1);
            \dataplot{XPOS}{Accuracy (XPOS)}{Part-of-Speech Tagging}
            \dataplot{SA}{macro-F1}{Sentiment Analysis}
        \end{groupplot}
        \coordinate (centre) at ($(column1)!.5!(column2)$);
        \node[below] at (centre |- current bounding box.south){\pgfplotslegendfromname{legend:size}};
    \end{tikzpicture}
    \caption{
        Downstream task performance as the pre-training data size grows.
        All values are the mean over 5 runs with different random seeds.
        The standard deviation being represented by the corresponding shaded area.
    }\label{figure:size}
\end{figure*}
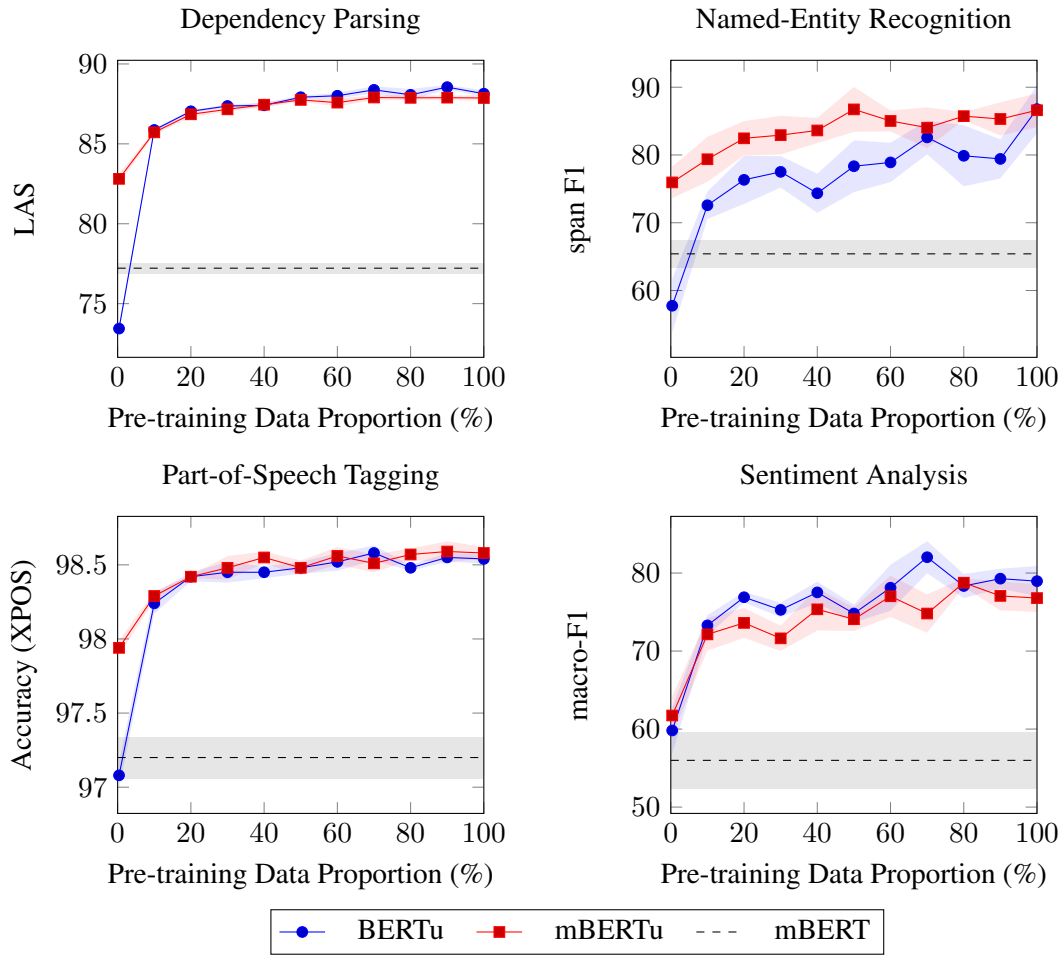

We do this by fixing the desired data proportion and scaling the pre-training data to satisfy this proportion, keeping the original training and validation split.
In tandem, the original 1M and 250K steps and batch size used in Section~\ref{section:models} are scaled down with the data size to pre-train for the same number of epochs as the models with the entire data.
Language models at 10\% intervals are pre-trained, with 100\% being the original models from Section~\ref{section:models}.
In this analysis, we also include the BERTu Mixed and mBERTu Mixed models from Section~\ref{section:analysis-domains}, which use 0.38\% of the data, estimated as a proportion of sentences.

After pre-training, each language model is fine-tuned on each of the downstream tasks in the same setup considered in Section~\ref{section:evaluation}.
These results are visualised in Figure~\ref{figure:size}.

As expected, the performance generally improves with  more pre-training data.
Surprisingly, the performance gap between the monolingual and multilingual models is drastically reduced with just 10\% of the data.
With this little data all configurations outperform mBERT.
For Named-Entity Recognition this is also the case but it takes around 70\% of the data for BERTu and mBERTu to start achieving very close performance.

It is also noticeable that the gradual increase is not monotonic, although it is more stable for Dependency Parsing and Part-of-Speech tagging.
Suprisingly, BERTu with 70\% of the data performs better than with 100\% of the data on Sentiment Analysis.
Similarly, mBERTu with 50\% of the data performs better than with 100\% of the data on Named-Entity Recognition.
One possible explanation may be due to the relationship between the number of steps and batch size chosen, but further investigation is warranted.

On Sentiment Analysis, BERTu is consistently better than mBERTu with 10\% or more of the data, and is at times significantly better.
This finding gives some evidence that monolingual representations seem better suited for fine-tuning on semantic tasks in a specific language.

\section{Conclusion}\label{section:conclusion}

In this work we analyse the impact of pre-training data on downstream task performance in a low-resource setting, specifically focusing on Maltese.
We present a newly developed corpus of around 500M tokens, which allows us to study how the pre-training data size and domain translates in downstream performance differences.
Using BERT as our architecture, we compare a monolingual language model, pre-trained from scratch, to a further pre-trained multilingual model, in a number of pre-training configurations.
We conduct an evaluation on a both syntactic and semantic tasks.

In line with previous findings on domain pre-training (\citealp{gururangan-etal-2020-dont}; inter alia), we find that matching the pre-training domain to the target task domain, results in improvements.
Moreover, we demonstrate that pre-training language models with varied domains is often beneficial over pre-training solely with Wikipedia.
These adjustments were in certain cases enough to surpass mBERT, underlining the importance of having pre-training data more suited to the target task, even at a small scale.

Whilst we show that further pre-training data does improve downstream performance, the gains are linear with exponential increases in data.
In fact, substantial improvements are observed with a small proportion of the pre-training data, over language models trained with Wikipedia-sized data.
This echoes the findings made by \citet{martin-etal-2020-camembert} with a small pre-training subset, although our reduced data setup is considerably smaller.

Using the whole corpus, we also pre-train two new language models: BERTu, a monolingual BERT model, trained from scratch, and mBERTu, which is the result of further pre-training mBERT.
These models demonstrate state-of-the-art results in Dependency Parsing, Part-of-Speech Tagging, Named-Entity Recognition, and Sentiment Analysis.
Moreover, we show that in general, BERTu performs better than mBERTu, as well as other baselines.
Through this, we also demonstrate that language-specific pre-training is most beneficial for higher-level tasks.

Despite these considerable improvements, the pre-training setups used in this work are as close as possible to the baselines, to allow for a more controlled comparison.
Hence, in the future, we plan to experiment with more language-specific tuning to push the state-of-the-art even further.

Even though this new corpus will undoubtedly improve the state of resources available for Maltese, the language is by no means a highly-resourced one.
The corpus we use is significantly smaller than typically used corpora for higher-resourced languages.
We also remark that the quantity of labelled data is still scarce, and at times non-existent for certain tasks.
Although we include a semantically-oriented task in our evaluation, future work should investigate the efficacy of these models in more complex Natural Language Understanding scenarios.

We make this corpus and the models publicly available to foster further work and improvements for various NLP applications for Maltese.
We also hope that this work inspires work in other low-resource languages, since we show that the amount of data needed to achieve considerable improvements, does not need to be overly ambitious.

\section*{Acknowledgements}
This work is partially funded by the Malta Digital Innovation Authority (MDIA) under the Malta AI Strategy Framework 2019.
We also acknowledge  LT-Bridge Project (GA 952194) and DFKI for access to the Virtual Laboratory.
We are also grateful to the University of Malta Libraries for granting access to their digital open repository and to the many Maltese authors who gave permission for their work to be included in the Korpus Malti v4.0.

\bibliography{anthology,custom}
\bibliographystyle{acl_natbib}

\appendix

\end{document}